# Improving a Credit Scoring Model by Incorporating Bank Statement Derived Features

Rory P. Bunker [1], M. Asif Naeem [2], Wenjun Zhang [3]

*Auckland University of Technology*

*55 Wellesley St E, Auckland 1010, New Zealand*

**Abstract**

In this paper, we investigate the extent to which features derived from bank statements provided by loan applicants, and which are not declared on an application form, can enhance a credit scoring model for a New Zealand lending company. Exploring the potential of such information to improve credit scoring models in this manner has not been studied previously. We construct a baseline model based solely on the existing scoring features obtained from the loan application form, and a second baseline model based solely on the new bank statement derived features. A combined feature model is then created by augmenting the application form features with the new bank statement derived features. Our experimental results show that a combined feature model performs better than both of the two baseline models, and that a number of the bank statement derived features have value in improving the credit scoring model. As is often the case in credit scoring, our target data was highly imbalanced, and Naive Bayes was found to be the best performing classifier, outperforming a number of other classifiers commonly used in credit scoring. Future experimentation with Naive Bayes on other highly imbalanced credit scoring data sets will help to confirm whether the classifier should be more commonly used in the credit scoring context.

[1]Corresponding Author: rorybunker@gmail.com
[2]mnaeem@aut.ac.nz
[3]wzhang@aut.ac.nz

## 1. Introduction

Credit scoring models are widely used by banks and other financial institutions, in order to assess the risk of default of applicants for loans. Credit scoring can be thought of as a classification problem. Credit scoring models take a vector of attributes for a loan applicant, and given these attributes, attempt to discriminate between goods and bads; that is, to discriminate between those that are not likely to default or be in arrears with their payments, and those that are.

A lending company in New Zealand scores loan applicants based on 11 attributes obtained from an application form, which is filled out online when applying for a loan. The lending company had recently begun to incorporate data from an application called Credit Sense[1] into their data warehouse, which automatically extracts line-by-line bank statement data for a 90-day period of spending. The bank statement data is currently used for purposes such as income verification. The company was interested in exploring the potential of this data for credit scoring purposes.

The company was firstly interested in seeing whether a sufficiently predictive scoring model could be created using only the bank statement derived features, in which case the questions in the online application form would then be unnecessary to ask. Loan applicants would then be able to complete the application process faster, and the lending company would potentially be able to write more loans. If this model was not found to be sufficiently predictive, then the lending company was interested in investigating the extent to which the bank statement derived features could be of use in developing an improved scoring model by supplementing existing scoring features. A number of machine learning and statistical methods have been used for experimentation in this paper to develop this new scoring model.

The novelty of our contribution is in the investigation of deriving features from bank statement data provided by loan applicants, which are not declared on the loan application form, to be used in credit scoring models. To the best of our knowledge, this has not been studied previously. For this investigation, firstly we construct **two baseline models**. The **first**, based solely on the existing scoring features, and the **second**, based solely on the new bank statement derived features. Secondly, we create a **combined**

---

[1] https://creditsense.com.au
2

**feature model** by combining the existing scoring features with the new bank statement derived features.

Our experimental results using ROC analysis show firstly that the application form feature baseline model outperforms the bank statement baseline model, and in turn, a combined feature model performs better than both of the two baseline models. This indicates that the bank statement derived features have value in supplementing, but not in replacing, the application form features. A Naive Bayes model consisting of the 11 existing application form features, along with 5 new bank statement derived features, was found to be the best performing model.

The rest of the paper is organized as follows. Section 2 provides a background on credit scoring. Section 3 presents our research goals and methodology – in particular, the knowledge discovery in databases framework that is used to guide the data mining and knowledge discovery process. Section 4 outlines the candidate models that are used in the experimentation, our experimental approach, and the results of the experiments. Section 5 provides some discussion of the results, and Section 6 concludes the paper and presents some avenues for further work.

## 2. Credit Scoring

Thomas et al. (2002) defined credit scoring as "the set of decision models and their underlying techniques that aid lenders in the granting of consumer credit. These techniques decide who will get credit, how much credit they should get, and what operational strategies will enhance the profitability of the borrowers to the lenders". The present study is focused on models deciding who will get credit (application scoring), rather than on modelling the behaviour of existing loans (behavioural scoring).

There has been a significant amount of academic research in the area of credit scoring. There exists a number of review papers in the literature on credit scoring including those by Rosenberg & Gleit (1994), Hand & Henley (1997) and Thomas (2000). More recent expositions include that by Abdou & Pointon (2011). Yu et al. (2008) also provided a useful summary in chapter 1 of their book, although the book primarily focused specifically on the use of Support Vector Machines for credit scoring.

There are a number of important benefits from making use of credit scoring models. While initially credit analysts in the industry used methods based on their own judgment, sophisticated statistical methods found their way into use and have generally been found to be more objective, as well as enabling the large scale automation of the loan acceptance process, and writing more loans than they otherwise would have been able to, at lower cost. Economic pressures from an increase in the demand for credit, as well as greater competition, and the development of new technology, are all factors that have led to the development of statistical models to aid the lending decision (Hand & Henley, 1997). Not only are scoring models the only way of handling the large number of applicants nowadays, but also they seemingly produce more accurate classifications than the subjective judgemental assessments by human experts (Rosenberg & Gleit, 1994).

In assessing the credit risk of a loan applicant, a lending company typically obtains applicant characteristics via an application form, and often from credit bureau checks, which check the past credit history of applicants. In some cases, an initial application form screens good and bad risk applicants, and only those who proceed past this stage are checked by credit bureau agencies. This sequential assessment of loan applicants avoids unnecessary costs and provides quick credit application decision time to most applicants (Hand & Henley, 1997).

As mentioned earlier, credit scoring can be viewed as a classification problem, where we attempt to discriminate between good and bad risks, given their characteristics. A useful mathematical summary of the traditional credit scoring evaluation problem was outlined in Yu et al. (2008), and is summarized below. Let

$$\mathbf{x} = (x_1, x_2, .., x_m)^T \qquad (1)$$

be a vector of $m$ random variables that describe the information from a customer's application form and credit bureau information. The actual value of the attributes for a specific loan applicant $k$ are

$$\mathbf{x}_k = (x_{1k}, x_{2k}, .., x_{mk})^T. \qquad (2)$$

All samples are then given by

$$S = (\mathbf{x}_k, y_k), \, k = 1, 2, .., N \qquad (3)$$

where $N$ is the number of loan applicant samples in the data set, $\mathbf{x}_k$ is the attribute vector of the $k$th applicant, and $y_k$ is the observed result of whether or not there was timely repayment. For example, we could define $y_k = 1$ ('good') if the customer repaid the loan on time, and $y_k$ = 0 ('bad') if applicant $k$ did not repay the loan (or was in arrears for some time).

In cases where we have both the attribute vector and the observed loan outcome, the data is labelled and thus we are able to make use of supervised learning models for classification from statistics and machine learning.

In order to make an accurate judgement of the likelihood of $y_k$, given the applicants attribute vector $\mathbf{x}_k$, there have been a number of models that have been applied to the problem, drawing from different disciplines including machine learning, statistics and operations research. By analysing the techniques used in a large number of research papers, Yu et al. (2008) grouped these techniques broadly to fit within 5 categories as shown in Table 1.

| Technique Category | Technique |
|---|---|
| Statistical Models | Linear Discriminant Analysis |
| | Logistic Regression |
| | Probit Regression |
| | K-Nearest Neighbours |
| | Decision Trees |
| Mathematical Programming | Linear Programming |
| | Quadratic Programming |
| | Integer Programming |
| Artificial Intelligence | Artificial Neural Networks (ANN) |
| | Support Vector Machines (SVM) |
| | Genetic Algorithm |
| | Genetic Programming |
| | Rough Set |
| Hybrid Approaches | ANN and Fuzzy Systems |
| | Rough Set and ANN |
| | Fuzzy System and SVM |
| Ensemble Approaches | ANN Ensemble |
| | SVM Ensemble |
| | Hybrid Ensemble |

Table 1: Techniques for Credit Scoring (Yu et al., 2008)

While traditionally statistical and linear programming techniques were most commonly applied in credit scoring, sophisticated methods from artificial intelligence and machine learning have also become commonly applied in the area of credit scoring.

## 3. Research Goals and Methodology

*3.1. Research Goals*

The primary goals of this research are as follows:

1. Extract relevant features from the bank statement data which is a non-trivial task. These features are then potentially used for modelling credit risk.
2. Constructing following baseline models:
   - based solely on the existing scoring feature set, obtained from the information declared on the loan application form.
   - based solely on the bank statement derived feature set, obtained from bank statement information provided by loan applicants via Credit Sense.
3. Comparing the performance of the combined feature model (both the application form and bank statement derived features) with the above defined two baseline models. The key purpose of the performance comparison is to investigate whether the bank statement derived features are sufficiently predictive by themselves for a credit scoring model, or the extent to which the bank statement derived features can supplement the existing scoring features to enhance a credit scoring model.

*3.2. Methodology*

For this project, a structured experimental approach to the data mining process was sought, in order to increase the likelihood of achieving good results. The knowledge discovery in databases process (KDP) outlines a formal experimental approach to extracting and creating knowledge from data. The KDP is defined as the non-trivial process of identifying valid, novel, potentially useful, and ultimately understandable patterns in data (Fayyad et al., 1996). The KDP involves having some input data, and in the end, coming out with some new knowledge that is valuable and useful to the organization.

KDP frameworks were initially developed in academia, and frameworks from industry followed (Cios et al., 2007). Two models that are considered to be leading ones are that developed by Fayyad et al. (1996), which was developed from academic research, and the second is the Cross Industry Standard Process for Data Mining (CRISP-DM) model (Chapman et al.,

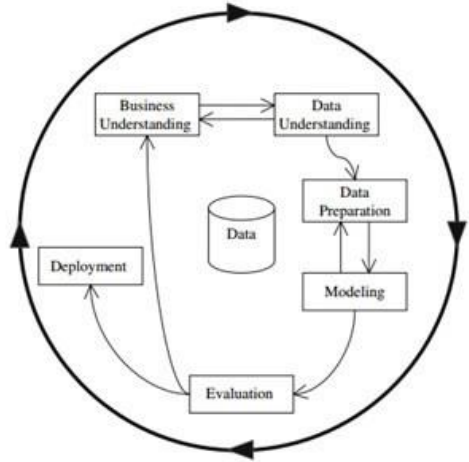

Figure 1: Steps in the CRISP-DM model of the KDP. 1. Business Understanding 2. Data Understanding 3. Data Preparation 4. Modeling 5. Evaluation 6. Deployment. Source: Cios et al. (2007).

2000), which was developed out of a project led by a number of organizations from industry.

Due to this project being conducted with industry, we decided to apply the CRISP-DM methodology. This framework consists of 6 main steps as shown in Figure 1.

In the first two steps, the data miner needs to gain understanding of the business or application domain. In the current project, this involved becoming familiar with the lending company and its data sources. This is a natural process of spending time in the company, talking with staff, as well as preliminary investigation and querying of the lending company's data warehouse.

Data preparation or preprocessing involves the (often time-consuming) process of creating a final target data set that can be used as an input to the machine learning or statistical model. The data preprocessing indeed proved to be a time consuming aspect of the present study, and involved extracting new features from the bank statement data, and then augmenting these features with the existing scoring features. Nonetheless, it is a very important step as it can considerably affect the success of the data mining experiments and helps to ensure that one does not obtain misleading results (Pyle, 1999).

Algorithm selection in the present study involves investigating previous literature in the application domain, and selecting and experimenting with algorithms that have been successful in previous studies. The experimentation with the different candidate models constitutes the data mining step.

The final steps are to evaluate the model, its potential use to the company, and potentially to deploy it in some way. This involves putting the results and/or knowledge obtained from the model to use e.g. through deploying the model into existing company IT systems, or in the production of a report outlining the analysis and results from the project.

## 4. Experiments

*4.1. Experimental Arrangements*
*4.1.1. Candidate Models*

An important step in the knowledge discovery process is selecting candidate models. The candidate models that were used for experimentation in this study are as follows:

1. Logistic Regression
2. Naive Bayes
3. Support Vector Machine (SVM) with a linear kernel
4. Nearest Neighbor
5. J48 Decision Tree
6. Random Forest

Linear Discriminant analysis (LDA) was not considered in this study due to its difficulty in dealing with nominal features. Neural Networks were not used because of their lengthy training time, and lack of interpret ability. We also decided not to consider mathematical programming approaches, as to consider these would require a completely different experimental set-up in terms of defining objective functions and constraints. Naive Bayes, interestingly, was not noted by Yu et al. (2008) as being a common technique in credit scoring; however for comprehensiveness, it was included in this study.

To simplify the experimentation, all learning schemes were run in WEKA (Hall et al., 2009) with their default parameter settings.

We now briefly outline some of the theoretical background of each of the candidate models that will be used in the experimentation.

Logistic Regression is a general linear model (GLM) that models a binary outcome (0/1, good/bad etc.) on a certain number of predictors. Logistic regression is the most commonly applied and a strong performing method for credit scoring. A strong theoretical basis in that is that it directly gives us an additive log odds score, which is a weighted linear sum of the attribute values (Thomas, 2009).

A logistic transformation yields a dependent variable that is between 0 and 1, and this provides an intuitive interpretation, being the probability of default given the attribute vector for that particular loan applicant. The logistic function is given by

$$p(x) = \frac{e^x}{1+e^x} = \frac{1}{1+e^{-x}}. \tag{4}$$

In logistic regression, we substitute the standard linear regression model into the logistic function (4). That is,

$$y = \beta \cdot \mathbf{x} = \beta_0 + \beta_1 x_1 + \beta_2 x_2 + \ldots + \beta_k x_k \tag{5}$$

$$p(\mathbf{x}) = \frac{1}{1+e^{\beta \cdot \mathbf{x}}} \tag{6}$$

where $y$ is the class value (good, bad), and $\mathbf{x}$ is a vector of applicant characteristics.

Rearranging (6) and taking the natural logarithm of both sides yields

$$\frac{p(\mathbf{x})}{1-p(\mathbf{x})} = e^{\beta_0 + \beta_1 x_1 + \beta_2 x_2 + \ldots + \beta_k x_k}, \tag{7}$$

$$\ln \frac{p(\mathbf{x})}{1-p(\mathbf{x})} = \beta_0 + \beta_1 x_1 + \beta_2 x_2 + \ldots + \beta_k x_k. \tag{8}$$

If $p(\mathbf{x})$ is defined as taking the value 0 if the loan applicant is a bad risk, and 1 if the loan applicant is a good risk, then the term $p(\mathbf{x})$ is the conditional probability of being a good risk, given the loan applicants attribute vector $\mathbf{x}$.

Estimation of the coefficients of the model in (8) can be achieved using maximum likelihood estimation (MLE), where we try to find estimates for the beta coefficients such that the predicted probability of default for each individual corresponds as closely as possible to the individuals observed status (James et al., 2013). This MLE procedure is built into most modern statistical computing packages.

The Naive Bayes classifier is a probabilistic classifier that uses the wellknown Bayes Theorem from probability theory. We follow the notation of Hall et al. (2011), and state Bayes Theorem as

$$P(H|E) = \frac{P(E|H)P(H)}{P(E)} \quad (9)$$

where $H$ denotes the hypothesis (the class of the particular instance), and $E$ represents the evidence (the attribute values for the particular instance). Given an applicant attribute vector **x** and a new instance to classify, if $P(Good|\mathbf{x})$ is larger than $P(Bad|\mathbf{x})$ then we classify the new instance as good, but otherwise we classify the new instance as bad.

The key assumption of Naive Bayes is that the attributes are conditionally independent of each other, given the class. When we have conditional independence of attributes given the class value, we can write (9) as

$$P(H|E) = \frac{P(E_1|H)P(E_2|H)...P(E_n|H)P(H)}{P(E)} \quad (10)$$

Where $P(H)$ is the prior probability of obtaining class $H$, regardless of the attributes, and the $P(E_i|H)$ denote the probability of obtaining that attribute value $E_i$, given the class value. Although we do not have the denominator in Bayes Theorem $P(E)$, one can normalize the likelihoods so that the $P(H|E)$ probabilities add up to one for the different class values.

Although the conditional independence assumption rarely holds in practice, the Naive Bayes classifier has nonetheless often provides impressive results (Hall et al., 2011). Its probabilistic foundation is also relatively interpretable and intuitive.

Support vector machines (SVM) perform classification by attempting to divide the points that belong to the two classes by constructing the largest possible 'wedge' between the two classes. The SVM was first proposed in Cortes & Vapnik (1995).

The SVM produces nonlinear boundaries by constructing a linear boundary in a large, transformed version of the feature space (by using a kernel function), and the method does not require that the classes be linearly separable (James et al., 2013). A soft margin allows some observations to be on the incorrect side of the margin or hyperplane. A parameter known as the complexity parameter (often denoted by $C$) reflects the number and seriousness of violations to the margin and hyperplane that we are willing to tolerate.

By using a soft margin, SVM usually generalizes well. SVM has proved to be a powerful method for classification that has outperformed other methods in a wide variety of application domains e.g. text categorization and face and fingerprint identification, and has been successful in various studies of credit risk evaluation (Yu et al., 2008). As a result of the SVM depending primarily on (possibly few) support vectors, the method is quite robust to data noise. A weakness of SVM is that the output of the model (weights) are not particularly interpretable.

The *k*-nearest neighbors method classifies an unseen data instance using the classification of the instance(s) that are closest to the unseen instance, using some specified distance metric (Bramer, 2007). The algorithm determines the distance between the test instance and every training example and selects the set of *k* training instances to the test example, then it predicts the most common class of these nearest training examples by a majority-voting scheme (Pang-Ning et al., 2006). In the special case where *k* = 1, we simply refer to the algorithm as the nearest neighbor classifier. For simplicity, we use *k* = 1 in the experiments in this paper. By far the most common distance metric used is Euclidean distance. The Euclidean distance *D* between two attribute vectors in *n*-dimensional Euclidean space **x** = ($x_1, x_2, ...x_n$) and **y** = ($y_1, y_2, ..., y_n$) is given by

$$D(x,y) = \sqrt{\sum_{i=1}^{n}(x_i - y_i)^2}.$$
(11)

An advantage of the nearest neighbor classifier is in its simplicity, in that the algorithm is relatively easy to understand and can guide one's intuition (Aha et al., 1991). Its theoretical basis is appealing when considering how the algorithm is aligned intuitively to how traditional loan decisions were made via human judgement, where credit officers compared the application characteristics to previous similar applicants in order to classify a new loan applicant. Another advantage is that since all work is done when we want to classify a new instance rather than when the training data is processed (Hall et al., 2011), classification speed is very fast. On the other hand, since the distance of all combinations between test and training instances must be computed, the training speed of the algorithm can be slow.

The J48 decision tree is the WEKA implementation of the C4.5 decision tree developed by Quinlan (1993). Decision Trees summarize training data in a tree structure, where each branch represents an association between attribute values and a class label (Novakovi´c et al., 2011). The tree recursively splits the training data based on tests on the potential of feature values in separating the classes (where for C4.5/J48, potential is measured by information gain). New test instances are simply classified by following the decision rules from the top node of the tree downwards.

The main advantage of the decision tree classifier is in its output: the tree structure can provide a compact and interpretable structure with decision rules to classify fresh test instances. However, if not pruned adequately, trees can become overly complex and can overfit the training data.

Random Forests (Breiman, 2001) are an ensemble method, which combine the votes of multiple decision trees that have been generated, through a bagging procedure. Candidate features to split on are selected randomly, from a set of a number of the most important features. The random forest calculates a response variable *y* by creating many (usually several hundred the default in WEKA is 100) different decision trees and then putting each object to be modeled down each of the decision trees - the response is then determined by evaluating the responses from all of the trees (Horning et al., 2010).

Although random forests have proven to usually perform better than single decision trees, Random Forests lose their interpretability when compared with a single decision tree structure.

*4.1.2. Model Evaluation Metrics*

A common way to evaluate the performance of a classifier is by observing a confusion matrix (Table 2). A confusion matrix shows all of the instances in the data set, categorized into four different categories: Goods that were classified correctly as such (TP/True Positives), Bads that were classified correctly as such (TN/True Negatives), Goods that were incorrectly classified as Bads (FN/False Negatives), and Bads that were incorrectly classified as Goods (FP/False Positives).

|  | **Classified Bad** | **Classified Good** |
|---|---|---|
| **Actual Bad** | TN | FP |
| **Actual Good** | FN | TP |

Table 2: Confusion Matrix in Credit Scoring

The true positive rate is the proportion of actual goods that were correctly classified as such:

$$TPRate = \frac{TP}{TP + FN}. \tag{12}$$

Similarly, the true negative rate is the proportion of actual bads that were classified correctly as such:

$$TNRate = \frac{TN}{TN + FP} = 1 - FPRate. \tag{13}$$

In the credit scoring context, the costs of misclassification are quite different. False negatives only refer to the opportunity cost of lost interest that could have been gained, whereas for false positives, the lender loses some or all of not only the interest, but also the principal that was to be repaid (Abdou & Pointon, 2011). Therefore, false positives are significantly costlier since these are people who were classified as being good by the model, and granted a loan, but actually turned out to be bad.

Accuracy is the most common metric to evaluate the performance of classifiers. Accuracy represents the proportion of total instances that were classified correctly by the model:

$$Accuracy = \frac{TP + TN}{TP + FP + TN + FN} \tag{14}$$

When the input data set is balanced, and there are equal costs associated with errors, accuracy is an appropriate measure to maximize when selecting a model. However, with imbalanced data and unequal error costs, the ROC curve or other similar techniques are more appropriate (Chawla et al., 2002). In the case of highly imbalanced data (as we have in our data set), it is trivial for a classification model to obtain very high accuracy by simply classifying every loan applicant as a good (Elkan, 2001). But if all loan applicants are classified as good by the model, no actual bad applicants will be correctly classified as bad, which would obviously be a poor model for our purposes.

An ROC curve plots the true positive (TP) rate – the proportion of good risks that are correctly classified, as a function of the false positive (FP) rate, for the whole possible range of cut-off values (Bastos, 2007). It is often easier to think of the x-axis as 1–*TNRate*. The ROC curve represents a trade-off between TP Rate, and TN Rate. Resampling or cost sensitive classification can move us to a different point on the ROC curve. More favorable models have ROC curves that are situated towards the top-left hand corner of the (TP, FP) space.

Rather than attempting to maximize accuracy, in the presence of imbalanced data it is common to attempt to maximize the area-under-ROC-curve (AUC) measure. The AUC measure can show the general performance of a model to another model, in one number.

*4.1.3. Setup*

In this section, we outline some of the data preprocessing steps undertaken, as well as the characteristics of the target data set.

*Feature Extraction*

Data preprocessing is an important and often time consuming part of any real-world data mining project.

In the present study, data preprocessing firstly involved feature extraction from the lending company's data warehouse. The applicant features from the online application form, as well as the loan status (current, in arrears, defaulted), were held in one particular table in the data warehouse. The Credit Sense (bank statement) data was held in other separate tables in the data warehouse. OLAP queries in SQL Server were used to extract the data, and the two sets of features were then joined using the Pandas library in Python.

Relevant features to extract from the bank statement data were decided on in consultation with two senior staff members at the lending company, both of whom have deep knowledge of the credit industry.

Location and Merchant information was also initially extracted from the bank statement transactions, but were not found to be useful for modelling purposes (although location information is useful for address verification). The final feature set is shown in Table 3.

*Class Values: Defining 'Bad'*

In credit scoring, we often refer to good risks or simply 'goods', and bad risks or simply 'bads'. It is important to be clear in our definition of what represents the good class and what represents the bad class, for the class values in the target data set.

Upon consultation with a number of staff within the lending company, it was decided to use the following as the definition of bad: *a bad is someone who has either been 90 days or more in arrears or has already defaulted*.

Note that the 90 days in arrears aspect of this definition is consistent with the definition of default under the Basel II Accord (Siddiqi, 2012).

*Feature Selection*

In machine learning, feature selection methods are generally divided into two groups: filter methods and wrapper methods.

Wrapper methods are scheme dependent, and evaluate different possible subsets of features from the original set (using some search method) to come up with the best subset for that particular classifier (Hall et al., 2011). Due to the fact that they evaluate many different subsets of features using the learning scheme, wrapper methods are computationally expensive and can sometimes result in over fitting. They were not used in this study because of their computational expense.

Filter methods on the other hand, look at the predictive power of attributes using a particular measure, in a manner that is independent of the classifier used. Yang & Pedersen (1997) and Forman (2003) conducted comparative studies on filter methods in the context of text classification, and found that Chi-squared ranking and information gain ranking are among the best methods of feature selection for classification purposes (Yang & Pedersen (1997), Forman (2003) as cited in Geng et al. (2007)).

The approach in this study for feature selection was to rank the features based on their Chi-squared statistic and information gain with respect to the class, vary the number of features that are included in the model, and investigate the AUC value at different numbers of features. It was found that information gain produced very similar rankings to the Chi-squared statistic, and therefore ultimately it was decided to use only the Chi-squared statistic as the means of ranking features. Attribute selection is performed within the filtered-classifier meta learner in WEKA, so that we do not introduce bias by performing feature selection before training the model.

*Discretizing Numeric Features*

Numeric variables that are used in credit scoring are usually discretized into bins. Discretizing continuous variables has a number of advantages that are mentioned in Siddiqi (2012), including that doing so offers an easier way to deal with outliers, makes it easy to understand the nature and strength of relationships, and enables one to develop insights into the behaviour of risk predictors.

We discretize numeric features using the Discretize filter within the filteredClassifier meta-learner within WEKA, before applying the attribute selection filter. The binning algorithm used is the entropy-based multi-interval discretization method proposed by Fayyad & Irani (1992).

*Class Imbalance*

Data sets used for credit scoring are usually imbalanced, since the number of bad risks is usually small relative to the number of good risks. Our target data set was highly imbalanced, with the proportion of bad instances only 1.6% of the total number of instances (7401 good instances, 121 bad instances).

There are a number of approaches to dealing with the class imbalance problem. One of them is using cost-sensitive learning (Elkan, 2001), where penalties are imposed for particular types of misclassification. However, this approach relies on knowing the exact cost of misclassification, which can be difficult to determine.

Another approach is to reduce the number of instances in the good class in some way (under-sampling), or to boost the number of instances in the bad class (oversampling). A commonly used approach to oversampling the minority class is called Synthetic Minority Over-Sampling Technique (SMOTE), proposed by Chawla et al. (2002), which creates synthetic instances for the minority class by taking each minority class instance and introducing synthetic instances along the line segments joining any/all of the *k* minority class nearest neighbors.

Brown & Mues (2012) investigated the performance of a number of credit scoring models on real world data sets with different levels of imbalance, and found that random forests performed well with data sets with large imbalance, but SVM with a linear kernel, QDA and C4.5 all performed poorly with imbalanced data. Marqu´es et al. (2013) reviewed the effectiveness of a number of oversampling techniques (e.g. SMOTE and newer extensions of SMOTE) compared with undersampling techniques, and found that the oversampling techniques outperformed the undersampling techniques in most cases, especially when logistic regression was used.

*Experimental Setup*

Most learning schemes output probabilities as part of the classification process. For example, by default, logistic regression classifies an applicant as good if the conditional probability of being good, given their attribute vector *x*, is greater than 0.5, and bad otherwise. That is, If *P*(*Good*|**x**) > 0.5 then classify as Good; else classify as Bad.

It was found in the experimentation that using the default cut-off value of 0.5 generally resulted in poor discrimination in the bad class (nearly all bad risks would be classified as good risks), except for the Naive Bayes classifiers, which provided good performance regardless.

The threshold selector selects a mid-point threshold on the probability output of a particular classifier. The midpoint threshold altered from the default of 0.5, and is set so that a given performance measure is optimized. In this study, the threshold selector was run to optimize the F1 measure, which balances classifier precision and recall. The F1 measure is defined as the harmonic mean between precision *p* and recall *r* (Pang-Ning et al., 2006):

$$F1 = \frac{2pr}{p+r}. \tag{15}$$

The harmonic mean tends to be closer to the smaller of the two values used in the calculation.

Threshold selection was performed with all classifiers for the purposes of initial comparison of candidate models. Once we have found the best performing model, we will then investigate the use of cost sensitive classification (rather than a resampling approach) to deal with the class imbalance.

The model configurations in WEKA were set up so that base classifier (Naive Bayes, Logistic Regression, SVM etc.) were wrapped within the filtered-classifier meta-learner (which performs attribute discretization and attribute selection), and the filtered-classifier was in turn wrapped within the threshold-selector meta-learner.

The features in each of our three feature sets (bank statement derived, application form derived, and combined) were ranked according to their Chisquared Statistic with respect to the class. The combined feature set rankings are shown below in Table 3. All features were initially included in the model, and the lowest ranked features were eliminated from the model one by one. At each number of features, each classifier was run, and the AUC value was recorded.

*4.2. Results*

*4.2.1. Baseline Model with Application Form Features*

The AUC for each model, for the specific number of application form features, is shown in Figure 2. Naive Bayes was found to be the best perform-ing classifier, followed by Logistic Regression. Random Forest and Nearest neighbors yielded similar performance to each other, but were both inferior to Naive Bayes and Logistic Regression. It is noticeable that both the SVM and J48 models perform very poorly with this highly imbalanced data set (consistent with the findings of Brown & Mues (2012)), with both classifiers being unable to classify any bad applicants correctly.

| Rank | Feature | Source | Chi-Square |
|---|---|---|---|
| 1 | Demographic Feature 1 | Application Form | 140.5529 |
| 2 | Demographic Feature 2 | Application Form | 106.753 |
| 3 | Demographic Feature 3 | Application Form | 85.0343 |
| 4 | Loan Feature 1 | Application Form | 82.0138 |
| 5 | Bank Statement Transaction Feature 1 | Bank Statement | 79.9832 |
| 6 | Employment Feature 1 | Application Form | 55.6844 |
| 7 | Bank Statement Transaction Feature 2 | Bank Statement | 41.9988 |
| 8 | Employment Feature 2 | Application Form | 37.638 |
| 9 | Demographic Feature 4 | Application Form | 36.2622 |
| 10 | Demographic Feature 5 | Application Form | 32.9058 |
| 11 | Demographic Feature 6 | Bank Statement | 30.476 |
| 12 | Demographic Feature 7 | Application Form | 25.2823 |
| 13 | Credit Card Type 1 | Bank Statement | 16.987 |
| 14 | Bank 1 | Bank Statement | 12.5209 |
| 15 | Demographic Feature 8 | Application Form | 11.7183 |
| 16 | Demographic Feature 9 | Application Form | 9.5036 |
| 17 | Credit Card Type 2 | Bank Statement | 5.6001 |
| 18 | Bank 2 | Bank Statement | 4.722 |
| 19 | Credit Card Type 3 | Bank Statement | 4.1162 |
| 20 | Credit Card Type 4 | Bank Statement | 1.5046 |
| 21 | Bank Statement Transaction Feature 3 | Bank Statement | 0.8387 |
| 22 | Bank 3 | Bank Statement | 0.7701 |
| 23 | Bank 4 | Bank Statement | 0.3769 |
| 24 | Benefit Type 1 | Bank Statement | 0.3768 |
| 25 | Bank 5 | Bank Statement | 0.2995 |
| 26 | Benefit Type 2 | Bank Statement | 0.1636 |
| 27 | Bank 6 | Bank Statement | 0.0187 |
| 28 | Benefit Type 3 | Bank Statement | 0.0163 |

| 29 | Benefit Type 4 | Bank Statement | 0.0163 |

Table 3: Ranking of Combined Feature Set by Chi-squared Statistic with respect to the class. Note that the features have been anonymised for confidentiality purposes, as requested by the lending company

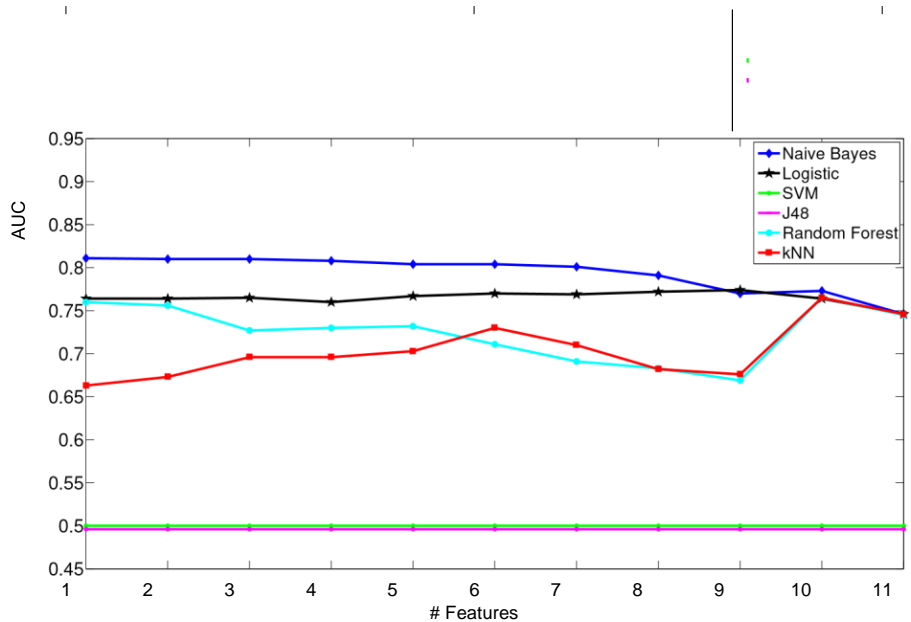

Figure 2: Comparison of AUC for each candidate model with the existing application form features

*4.2.2. Baseline Model with Bank Statement Derived Features*

The AUC for each model, for the specific number of bank statement derived features, is shown in Figure 3. Naive Bayes and Logistic Regression are the best performing classifiers with the bank statement derived features. Random Forest and nearest neighbor classifier again both had similar performance to each other. Comparing Figure 2 and Figure 3, we can see that AUC values for the candidate models are clearly higher with the application form feature set than with the bank statement derived feature set. Thus, we can conclude that the lending company should not develop a credit scoring model solely with the bank statement-derived features.

### 4.2.3. Combined Feature Set Model

We have established that models performed better with the application form feature set than with the bank statement derived feature set. We now investigate the extent to which the bank statement derived features can complement the existing application form features to improve AUC. The results using the combined feature set are shown in Figure 4. We can see that Naive Bayes is clearly the best performing classifier on the combined feature set. The Naive Bayes classifier reaches its highest AUC value (0.829) when 16 features are included in the model. Evidently, this feature set consists of all of the 11 existing application form features used by the lending company, along with 5 of the new bank statement derived features.

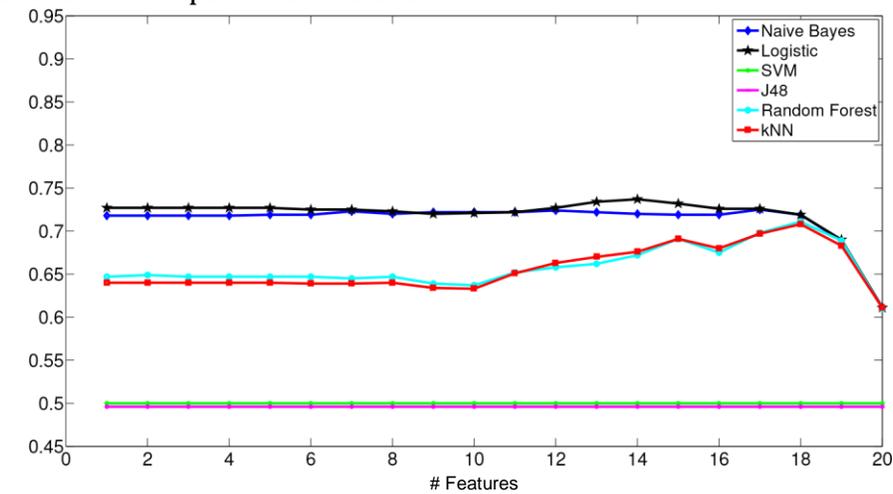

Figure 3: Comparison of AUC for each candidate model with the bank statement-derived features

## 5. Discussion

## 5.1. Evaluating the Final Model

We now investigate specifically whether the combined 16-feature

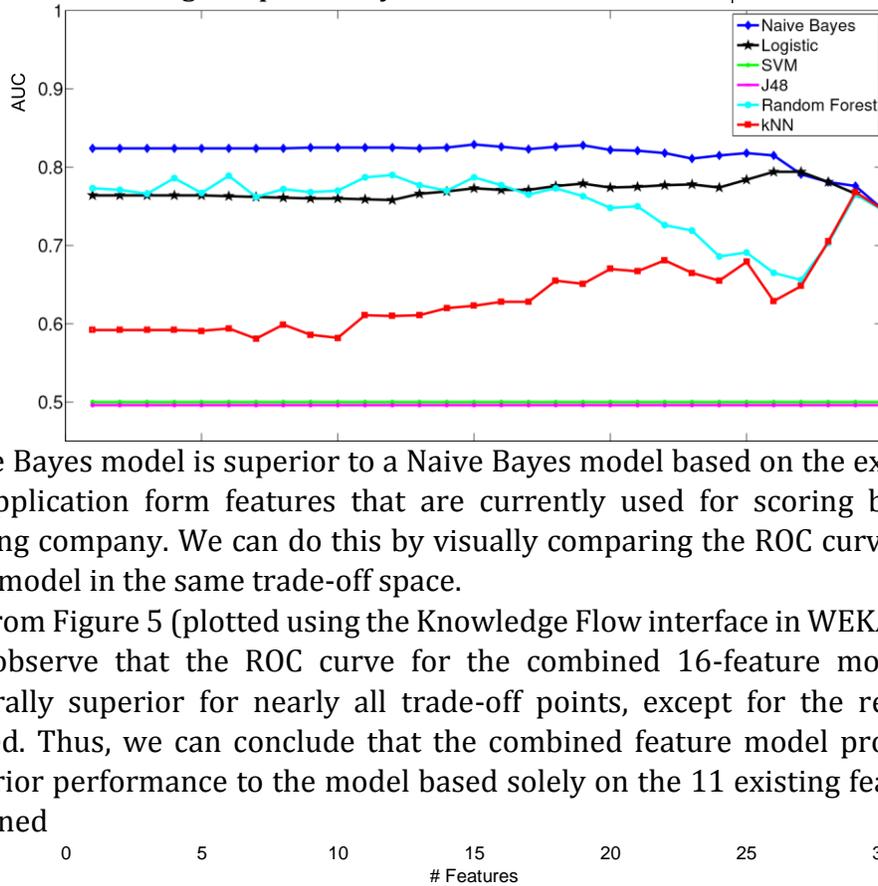

Naive Bayes model is superior to a Naive Bayes model based on the existing 11 application form features that are currently used for scoring by the lending company. We can do this by visually comparing the ROC curves for each model in the same trade-off space.

From Figure 5 (plotted using the Knowledge Flow interface in WEKA), we can observe that the ROC curve for the combined 16-feature model is generally superior for nearly all trade-off points, except for the regions circled. Thus, we can conclude that the combined feature model provides superior performance to the model based solely on the 11 existing features obtained from the application form, and therefore that the new features from the bank statements have improved the model.

Figure 4: Comparison of AUC for each candidate model with the combined feature set

## 5.2. Cost Sensitive Classification

A cost matrix contains the cost of misclassifying a good as a bad (false negative), and the cost of misclassifying a bad as a good. Estimation of these specific costs was not undertaken in the present study, but in any case, we can suppose that the cost of misclassifying a good as a bad is $1, and the cost of misclassifying a bad as a good $x$; in which case, the cost of misclassifying a bad as a good is $x$ times as great as misclassifying a good as a bad.

Table 4 shows the accuracy, TP rate, the number of goods out of the 7401 total that were classified correctly, the TN rate, and the number of bads out of the 121 total that were classified correctly by the combined 16-feature Naive Bayes model, for six different cost ratios. Note that as we increase the value of $x$, we can gain in true negative rate and move along the ROC curve, but at the expense of TP rate and overall accuracy.

Further work in conjunction with the lending company is needed to determine the exact costs of misclassification, and therefore to end up on the appropriate trade-off point on the ROC curve.

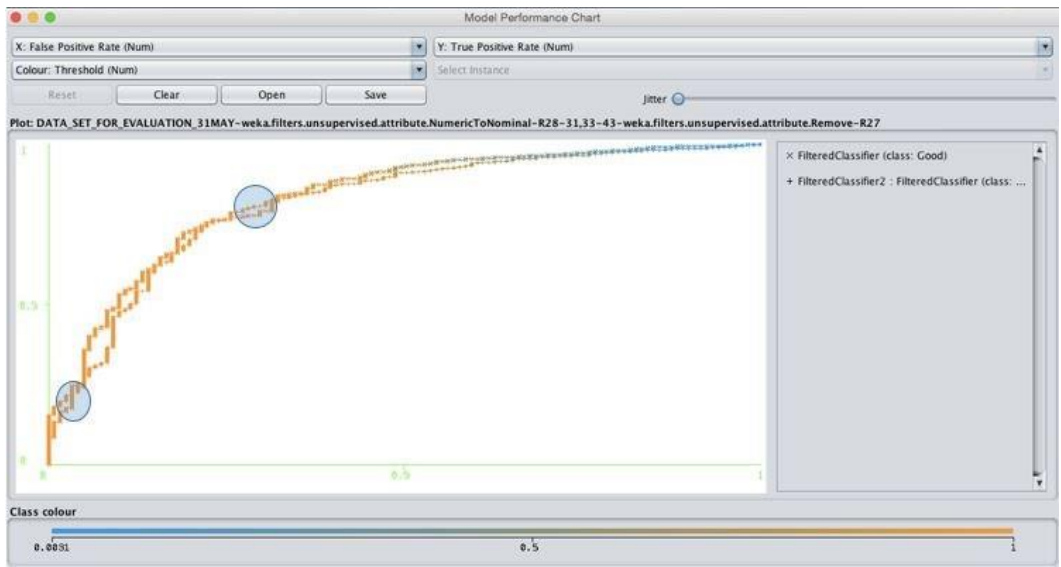

Figure 5: Naive Bayes for the combined feature model vs. Naive Bayes for the traditional feature model.

|                                              | 1     | 2     | 5     | 10    | 20    | 50    |
|----------------------------------------------|-------|-------|-------|-------|-------|-------|
| **Accuracy**                                 | 0.944 | 0.928 | 0.899 | 0.872 | 0.839 | 0.791 |
| **True Positive Rate**                       | 0.953 | 0.936 | 0.904 | 0.876 | 0.842 | 0.793 |
| **Goods Classified Correctly (out of 7401)** | 7059  | 6931  | 6700  | 6493  | 6239  | 5876  |
| **True Negative Rate**                       | 0.413 | 0.479 | 0.554 | 0.603 | 0.661 | 0.686 |
| **Bads Classified Correctly (out of 121)**   | 50    | 58    | 67    | 73    | 80    | 83    |

Table 4: 16-feature Naive Bayes Model Performance Metrics with different misclassification cost ratios. For example, the column with $x$ = 2 assumes that a false positive is twice as costly as a false negative.

### 5.3. Bank Statement Derived Features

As has been mentioned, in the final model, there were 5 features that were derived from the bank statement data, which were found to be of use in complementing the existing 11 application form features. Table 5 shows each of these features, bins and the weight of evidence (WOE) value for each bin. Note that the weight of evidence represents the natural logarithm of the odds ratio of the distribution of good relative to bad within that attribute bin (Siddiqi, 2012).

Since the features and their bins have been anonymised for confidentiality reasons, our full interpretation of the results in table 5 cannot be provided here. However, we can note that holding a credit card of type 1 is related to being a good risk, as well as not banking with bank 1, which is perhaps a reflection of customer quality of customers utilising these institutions.

| Attribute | Bin | WOE |
|---|---|---|
| Bank with Bank 1 | No | 0.159 |
|  | Yes | -0.568 |
| Hold Credit Card Type 1 | No | -0.267 |
|  | Yes | 0.585 |
| Demographic Feature 6 | low | -0.399 |
|  | high | 0.703 |
| Bank Statement Transaction Feature 2 | low | -0.514 |
|  | high | 0.719 |
| Bank Statement Transaction Feature 1 | low | 0.412 |
|  | high | -1.118 |

Table 5: Weight of evidence values for each of the 5 bank statement-derived attribute bins, in the final Naive Bayes combined feature model. Numeric features were discretized using the Fayyad & Irani (1992) method. The actual feature names and bin values have been anonymised for confidentiality purposes, as requested by the lending company.

## 6. Conclusions and Future Work

This research has supplemented features declared by loan applicants on a loan application form, that were currently used for credit scoring by a New Zealand lending company, with undeclared features that were extracted from bank statements provided by loan applicants during the loan application process. This work has highlighted the potential of using bank statement information from applications such as Credit Sense, to obtain additional features to improve credit scoring.

Although the bank statement features were not sufficiently predictive by themselves, our experimental results showed that a number of the bank statement derived features have value in supplementing the application form features, in improving a credit scoring model.

In particular, we found that a Naive Bayes model that used 16 combined features provided the best performance in terms of area under the ROC curve (AUC). This 16 feature model consisted of all 11 application form features that are currently used for credit scoring by the lending company, along with 5 new features derived from the bank statements.

There are a number of possible avenues for further work. One area is to extend this work is by incorporating reject inference methods. This involves attempting to assign a "value" to rejected applicants or determining the status of each reject and then using this "value" in the target data set (Thomas, 2009). This was not possible in the present study due to the non-availability of this type of data within the company.

This work has also highlighted the potential of Naive Bayes to be used with highly imbalanced data sets. It is interesting to note that Naive Bayes was not considered to be a commonly applied model in the credit scoring domain by Yu et al. (2008) (Naive Bayes is not in Table 1). We intend to perform further investigation of the ability of Naive Bayes to be used in the credit scoring context, and specifically with highly imbalanced data sets.

Another extension would be to further investigate the costs of misclassification with the lending company. This would involve coming up with accurate estimates of the cost of false negatives, and the cost of false positives. In doing so, the lending company would be moving to the appropriate trade-off point on the ROC curve.

From an organizational perspective, further work would be required in implementing the credit scoring model in practice e.g. integrating the model with the data warehouse and other IT systems, testing and so on.

In a separate line, data from bank statements obtained through applications like Credit Sense could be used in investigating possible fraud cases, for example, by using methods of outlier detection.

Elkan, C. (2001). The foundations of cost-sensitive learning. In *International joint conference on artificial intelligence* (pp. 973–978). LAWRENCE ERLBAUM ASSOCIATES LTD volume 17.

Fayyad, U., Piatetsky-Shapiro, G., & Smyth, P. (1996). From data mining to knowledge discovery in databases. *AI magazine*, *17*, 37.

Fayyad, U. M., & Irani, K. B. (1992). On the handling of continuous-valued attributes in decision tree generation. *Machine learning*, *8*, 87–102.

Forman, G. (2003). An extensive empirical study of feature selection metrics for text classification. *Journal of machine learning research*, *3*, 1289–1305.

Geng, X., Liu, T.-Y., Qin, T., & Li, H. (2007). Feature selection for ranking. In *Proceedings of the 30th annual international ACM SIGIR conference on Research and development in information retrieval* (pp. 407–414). ACM.

Hall, M., Frank, E., Holmes, G., Pfahringer, B., Reutemann, P., & Witten, I. H. (2009). The weka data mining software: an update. *ACM SIGKDD explorations newsletter*, *11*, 10–18.

Hall, M., Witten, I., & Frank, E. (2011). Data mining: Practical machine learning tools and techniques. *Kaufmann, Burlington*, .

Hand, D. J., & Henley, W. E. (1997). Statistical classification methods in consumer credit scoring: a review. *Journal of the Royal Statistical Society: Series A (Statistics in Society)*, *160*, 523–541.

Horning, N. et al. (2010). Random forests: An algorithm for image classification and generation of continuous fields data sets. In *Proceedings of the International Conference on Geoinformatics for Spatial Infrastructure Development in Earth and Allied Sciences, Osaka, Japan*. volume 911.

James, G., Witten, D., Hastie, T., & Tibshirani, R. (2013). *An introduction to statistical learning* volume 6. Springer.

Marqu´es, A. I., Garc´ıa, V., & Sa´nchez, J. S. (2013). On the suitability of resampling techniques for the class imbalance problem in credit scoring. *Journal of the Operational Research Society*, *64*, 1060–1070.Elkan, C. (2001). The foundations of cost-sensitive learning. In *International joint conference on artificial intelligence* (pp. 973–978). LAWRENCE ERLBAUM ASSOCIATES LTD volume 17.

Fayyad, U., Piatetsky-Shapiro, G., & Smyth, P. (1996). From data mining to knowledge discovery in databases. *AI magazine*, *17*, 37.

Fayyad, U. M., & Irani, K. B. (1992). On the handling of continuous-valued attributes in decision tree generation. *Machine learning*, *8*, 87–102.

Forman, G. (2003). An extensive empirical study of feature selection metrics for text classification. *Journal of machine learning research*, *3*, 1289–1305.

Geng, X., Liu, T.-Y., Qin, T., & Li, H. (2007). Feature selection for ranking. In *Proceedings of the 30th annual international ACM SIGIR conference on Research and development in information retrieval* (pp. 407–414). ACM.

Hall, M., Frank, E., Holmes, G., Pfahringer, B., Reutemann, P., & Witten, I. H. (2009). The weka data mining software: an update. *ACM SIGKDD explorations newsletter*, *11*, 10–18.

Hall, M., Witten, I., & Frank, E. (2011). Data mining: Practical machine learning tools and techniques. *Kaufmann, Burlington*, .

Hand, D. J., & Henley, W. E. (1997). Statistical classification methods in consumer credit scoring: a review. *Journal of the Royal Statistical Society: Series A (Statistics in Society)*, *160*, 523–541.

Horning, N. et al. (2010). Random forests: An algorithm for image classification and generation of continuous fields data sets. In *Proceedings of the International Conference on Geoinformatics for Spatial Infrastructure Development in Earth and Allied Sciences, Osaka, Japan*. volume 911.

James, G., Witten, D., Hastie, T., & Tibshirani, R. (2013). *An introduction to statistical learning* volume 6. Springer.

Marqu´es, A. I., Garc´ıa, V., & Sa´nchez, J. S. (2013). On the suitability of resampling techniques for the class imbalance problem in credit scoring. *Journal of the Operational Research Society*, *64*, 1060–1070.